\documentclass[
]{ceurart}

\usepackage{graphicx}
\usepackage{amsmath}
\usepackage{amssymb}
\usepackage{adjustbox}
\usepackage{booktabs}
\usepackage{float}
\usepackage{listings}
\usepackage{svg}
\usepackage{tikz}
\usepackage{xcolor}
\usetikzlibrary{positioning, arrows.meta, shapes.geometric}
\begin{document}

\copyrightyear{2026}
\copyrightclause{Copyright for this paper by its authors.
  Use permitted under Creative Commons License Attribution 4.0
  International (CC BY 4.0).}

\conference{CLEF 2026 Working Notes, 21 -- 24 September 2026, Jena, Germany}

\title{AI Wizards at EXIST 2026: Hierarchical Soft-Label Learning for Multimodal Sexism Identification in Memes}
\title[mode=sub]{Notebook for the EXIST Lab at CLEF 2026}

\author[1]{Matteo Fasulo}[%
orcid=0000-0002-7019-3157,
email=matteo.fasulo@sdsc.ethz.ch,
url=https://matteofasulo.com,
]
\cormark[1]
\fnmark[1]
\address[1]{Swiss Data Science Center, ETH Zürich, Andreasturm, Andreasstrasse 5, 8092 Zürich, Switzerland}

\author[2]{Antonio Gravina}[%
orcid=0009-0007-9252-263X,
email=antonio.gravina@everest-erp.com,
url=https://github.com/GravAnt,
]
\fnmark[1]
\address[2]{Everest Systems GmbH, Max-Jarecki-Straße 21, 69115 Heidelberg, Germany }

\author[3]{Luca Tedeschini}[%
orcid=0009-0006-0375-829X,
email=ltedeschini@villanova.ai,
url=https://github.com/LucaTedeschini,
]
\fnmark[1]
\address[3]{Villanova.ai S.P.A, Località Sa Illetta, SS 195 KM 2.3, 09123 Cagliari, Italy}

\author[4]{Luca Babboni}[%
orcid=0009-0001-5260-7467,
email=luca.babboni2@studio.unibo.it,
url=https://github.com/ElektroDuck,
]
\fnmark[1]
\address[4]{Independent researcher}

\cortext[1]{Corresponding author.}
\fntext[1]{These authors contributed equally.}
\begin{abstract}
We present the \textbf{AI Wizards} submission to EXIST 2026 for multimodal sexism identification in memes. The task is composed of three, increasingly harder subtasks. We model them hierarchically as conditional soft-label prediction over empirical annotator distributions. Our system maps fixed Gemini Embedding~2 vision-language representations through a lightweight Gated MLP trained with KL divergence and homoscedastic uncertainty weighting. Our submissions ranked \textbf{first} on Task~2.3 and \textbf{fourth} on Tasks~2.1 and~2.2 on the official Soft-Soft leaderboards. The code is available at \url{https://github.com/NLP-AI-Wizards/EXIST-2026}.
\end{abstract}


\maketitle

\section{Introduction}

Sexist content online reinforces women's marginalization and exclusion from digital spaces~\cite{nadim2021silencing}. Digital platforms often amplify offline patriarchal structures, where hostile and benevolent sexism jointly legitimize gender inequality~\cite{glick1996ambivalent}. Exposure to such content causes measurable psychological harm (e.g., elevated anxiety, depression, and self-censorship), which silences women's participation in public discourse~\cite{im2022womens}. Robust automated detection is therefore a critical intervention for digital safety.

Memes pose unique challenges: their meaning emerges non-linearly from text-image interaction, often relying on irony, humor, or implicit cultural knowledge to provide plausible deniability~\cite{fersini2022semeval, kiela2020hateful}. Detection systems must model how modalities jointly construct gendered meanings that may be implicit, ironic, or ambiguous.

EXIST 2026~\cite{plaza2026overview,plaza2026overview-extended} structures sexism detection 
as three hierarchically nested subtasks, detailed in Section~\ref{sec:dataset}: binary sexism 
identification, source intention detection for sexist content, and fine-grained multi-label 
categorization across five sexism types. Critically, the task adopts the Learning with 
Disagreement (LeWiDi~\cite{10.1613/jair.1.12752}) paradigm, requiring systems to predict the full empirical annotator distribution rather than a single majority label.

We address all three subtasks as hierarchical conditional multi-task learning over pre-computed vision-language embeddings, preserving semantic dependencies while directly optimizing for soft-label evaluation.

Our main contributions are as follows.
\begin{itemize}
    \item \textbf{Multimodal representation:} Pre-trained Gemini Embedding~2 features processed through compact gated MLP blocks, avoiding fine-tuning of large generative models on limited data.
    \item \textbf{Dynamic multi-task weighting:} KL-divergence soft-label objective with learned homoscedastic uncertainty weights that automatically balance task contributions.
    \item \textbf{Hierarchical decoding:} Conditional loss masking and probabilistic decoding that respect the task taxonomy without architectural overhead.
\end{itemize}

\section{Related Work}

\subsection{Multimodal Sexism Detection}
Early detection systems focused on text-only platforms~\cite{plaza2025overview, kirk2023semeval}. The introduction of multimodal benchmarks (SemEval-2022 MAMI~\cite{fersini2022semeval} and the Hateful Memes Challenge~\cite{kiela2020hateful}) revealed that unimodal systems fail when toxicity arises only from text-image interaction~\cite{cao2023procap}. The EXIST lab has tracked this evolution from text (2021–2023) to memes (2024)~\cite{plaza2024overview} to video (2025)~\cite{plaza2025overview}. As sexism detection has moved into the multimodal setting, it has increasingly drawn on architectures originally developed for the broader hate speech domain, which we review next.
\subsection{Architectures and Representation}
Contrastive vision-language models such as CLIP~\cite{clip} and SigLIP~\cite{siglip} learn powerful joint image-text representations, but their training objective is general-purpose and not tailored to hateful content detection. Building on this foundation, Hate-CLIPper~\cite{kumar-nandakumar-2022-hate} extends the contrastive paradigm by introducing a bilinear fusion mechanism explicitly designed for hate speech detection, effectively specializing the general alignment learned by CLIP-like models into a task-specific framework for identifying hateful memes. In parallel, an alternative line of work has moved away from fusion-based specialization altogether, instead leveraging the emergent reasoning capabilities of large pretrained models: recent EXIST submissions explored LoRA fine-tuning and prompting of large models such as Claude Sonnet~\cite{trippas2025mario} to tackle hateful meme detection directly, trading architectural specialization for scale and generality. These two directions, however, share a common limitation for the sexism detection setting: neither explicitly models the annotation disagreement that characterizes subjective tasks like sexism labeling, nor the hierarchical structure of sexism sub-categories.
\subsection{Learning with Disagreement and Hierarchy}
Subjective tasks such as sexism detection exhibit high inter-annotator disagreement that majority voting erases, potentially silencing minority perspectives. The LeWiDi paradigm treats this disagreement as signal rather than noise, evaluating systems on their ability to predict full annotator distributions. Hierarchical classification for related tasks has been explored through cascaded architectures and conditional decoding, but few systems combine both hierarchy and soft-label prediction in a unified multi-task framework.
Building on these three lines of work, our approach combines a compact contrastive backbone, following the efficiency principles of fusion-based architectures like Hate-CLIPper while avoiding the computational cost of large prompted models, with a disagreement-aware, hierarchical multi-task head that jointly predicts soft label distributions and sexism sub-categories. This design directly addresses the gap identified above: unlike prior CLIP-based or LLM-prompted systems, our model explicitly captures both annotator disagreement and hierarchical task structure within a single, efficient, and reproducible framework.

\section{Dataset and Problem Formulation}
\label{sec:dataset}

EXIST 2026 formalizes the research problem through three hierarchically organized subtasks focused on multimodal sexist content:
\begin{itemize}
    \item \textbf{Sexism Identification (Task 2.1):} A binary classification task aimed at determining whether a meme depicts a sexist situation or endorses sexist behavior (\texttt{YES}) or, conversely, does not contain sexist content (\texttt{NO}).
    \item \textbf{Source Intention Detection (Task 2.2):} A binary classification task defined exclusively for memes previously labeled as sexist. It distinguishes between \texttt{DIRECT} intention, where the communicative goal is to produce inherently sexist content or to incite sexist attitudes, and \texttt{JUDGEMENTAL} intention, where the content portrays sexist situations or behaviors with the explicit objective of condemning or criticizing them.
    \item \textbf{Sexism Categorization (Task 2.3):} A multi-label classification task that assigns each sexist meme to one or more of the following five categories:
    \begin{enumerate}
        \item \texttt{IDEOLOGICAL-INEQUALITY}: The content delegitimizes the feminist movement, rejects or denies structural inequality between men and women, or portrays men as victims of gender-based oppression.
        \item \texttt{STEREOTYPING-DOMINANCE}: The content conveys essentialist or stereotypical beliefs about women that imply they are more suited to specific roles (e.g., mother, wife, family caregiver, faithful, tender, loving, submissive), are unfit for certain activities (e.g., driving, physically demanding work), or explicitly or implicitly asserts male superiority.
        \item \texttt{OBJECTIFICATION}: The content depicts women as objects, disregarding their dignity and personhood, or prescribes or describes physical attributes that women are expected to possess in order to conform to traditional gender roles (e.g., adherence to narrow beauty standards, hypersexualization of female bodies, representation of women’s bodies as available for male use).
        \item \texttt{SEXUAL-VIOLENCE}: The content contains sexual insinuations, requests for sexual favors, or other forms of sexual harassment, including references to rape or sexual assault.
        \item \texttt{MISOGYNY-NON-SEXUAL-VIOLENCE}: The content expresses hatred, hostility, or non-sexual forms of violence directed toward women.
    \end{enumerate}
\end{itemize}

The dataset contains 3,984 training and 1,053 test memes in English and Spanish (Table~\ref{tab:dataset_size}). For the Soft-Soft track, each target represents the empirical annotator distribution. Tasks~2.2 and~2.3 are only semantically defined when sexism probability is non-zero, motivating our hierarchical conditional approach.

\begin{table}[htpb]
    \centering
    \caption{EXIST 2026 dataset composition.}
    \begin{tabular}{lccc}
        \toprule
        Language & Training & Test & Total \\
        \midrule
        Spanish & 1979 & 540 & 2519 \\
        English & 2005 & 513 & 2518 \\
        \midrule
        Total & 3984 & 1053 & 5037 \\
        \bottomrule
    \end{tabular}
    \label{tab:dataset_size}
\end{table}

\subsection{Physiological Data and Modality Selection}
\label{sec:physio}

EXIST 2026 provides auxiliary physiological recordings: EEG (16 channels, 5 frequency bands, 80 bandpower features), gaze data (200~Hz), and heart rate.

Among the three modalities, we prioritized EEG for quantitative analysis. EEG offers the highest feature dimensionality and the most direct, well-established link to implicit cognitive and evaluative processing in the psychophysiology literature~\cite{10.1093/scan/nsp004}, making it the most plausible candidate for capturing task-relevant variation among the provided physiological signals. Gaze and heart rate, by contrast, are lower-dimensional (respectively a 2D fixation trace and a single scalar signal) and reflect attentional and autonomic responses that are more directly tied to visual salience and general arousal than to the specific cognitive evaluation of sexist content. We therefore did not subject gaze and heart rate to the same statistical testing procedure; their potential utility, particularly in combination with EEG or under non-linear modeling, remains untested and is left for future work.

To assess their discriminative utility, we applied the Principal Component Analysis (PCA) to retain 21 components explaining $\sim95$\% of total variance to the EEG features, then performed Multivariate Analysis of Variance (MANOVA) in the majority groups of soft-labels for each subtask (Table~\ref{tab:manova_results}). Tasks~2.1 and~2.3 showed no significant linear separability. Although Task~2.2 reached statistical significance ($p_{\mathrm{perm}} = 0.0010$), the effect was negligible (Wilks' $\lambda = 0.9946$, partial $\eta^2 < 0.01$). No individual EEG channel survived the Bonferroni correction ($\alpha = 6.25 \times 10^{-4}$).

\begin{table}[htpb]
    \centering
    \caption{EEG separability analysis. PCA reduced features to 21 components. $p_{\mathrm{perm}}$ from 1,000-label permutation test.}
    \begin{tabular}{lccccc}
        \toprule
        Task & Sessions & Wilks' $\lambda$ & $p$ & $p_{\mathrm{perm}}$ & Sig.\ channels \\
        \midrule
        Task 2.1 & 15,477 & 0.9988 & 0.6130 & 0.6034 & 0/80 \\
        Task 2.2 & 10,171 & 0.9946 & $7.9 \times 10^{-5}$ & 0.0010 & 0/80 \\
        Task 2.3 & 10,171 & 0.9918 & 0.4851 & 0.4855 & 0/80 \\
        \bottomrule
    \end{tabular}
    \label{tab:manova_results}
\end{table}

Although non-linear biosignal integration remains an open direction, linear analysis revealed insufficient signal for the provided feature representations. We therefore prioritized robust semantic representations over physiological fusion, leaving deeper biosignal modeling for future work.

\section{Methods}

Our system rests on three principles: (1) extracting rich multimodal semantics without fine-tuning massive models on limited data; (2) predicting annotator disagreement distributions rather than forced labels; and (3) enforcing hierarchical constraints through conditional training and decoding.

\subsection{Shared Backbone}

Let $f_{\mathrm{VLM}}: \mathcal{X} \rightarrow \mathbb{R}^{768}$ denote the frozen Gemini Embedding~2 encoder~\cite{shanbhogue2026geminiembedding2native}, producing a fixed embedding $\mathbf{x}$ per meme. We pass $\mathbf{x}$ through a lightweight Gated MLP backbone consisting of one SwiGLU block~\cite{shazeer2020gluvariantsimprovetransformer} with expansion factor 2 and dropout 0.2:

\begin{equation}
    \mathrm{SwiGLU}(\mathbf{x}) = (\mathbf{x}\mathbf{W}_1) \odot \mathrm{SiLU}(\mathbf{x}\mathbf{W}_2),
\end{equation}
\begin{equation}
    \mathbf{h}^{(l+1)} = \mathbf{h}^{(l)} + \mathbf{W}_3\left(\mathrm{SwiGLU}\left(\mathrm{LN}(\mathbf{h}^{(l)})\right)\right).
\end{equation}

The gating mechanism learns to suppress benign confounders and amplify hostile cross-modal contradictions. The resulting representation $\mathbf{h}$ feeds three independent linear heads producing logits for each subtask. The trainable portion contains only 3.5M parameters.

\begin{figure}[htpb]
\centering
\begin{adjustbox}{max width=\linewidth}
\begin{tikzpicture}[
    input_node/.style={
        draw=blue!70!black,
        rectangle,
        rounded corners,
        minimum width=6cm,
        minimum height=1cm,
        align=center,
        thick,
        fill=blue!5
    },
    mlp_node/.style={
        draw=orange!80!black,
        rectangle,
        rounded corners,
        minimum width=6cm,
        minimum height=1cm,
        align=center,
        thick,
        fill=orange!10
    },
    var_node/.style={
        draw=teal!80!black,
        ellipse,
        text width=4.8cm, 
        align=center,
        thick,
        fill=teal!5,
        inner sep=2pt
    },
    op_node/.style={
        draw=black,
        circle,
        inner sep=1pt,
        thick,
        fill=gray!10,
        font=\large
    },
    >=Stealth,
    thick
]

\node[input_node] (gemini) {\textbf{Pre-trained Vision-Language}\\ Gemini Embedding 2 ($\mathbf{e} \in \mathbb{R}^{768}$)};

\node[mlp_node] (mlp) [below=1.2cm of gemini] {\textbf{Shared Projection Backbone}\\ (Stacked SwiGLU Expansion Blocks)};

\node[var_node] (t21) [below=1.5cm of mlp] {\textbf{Task 2.1}\\ $\hat{p}^{(1)} = P(\text{Sexist} \mid \mathbf{x})$};

\node[var_node] (t22) [below left=2.2cm and 0.5cm of t21] {\textbf{Task 2.2}\\ $\hat{p}^{(2)} = P(\text{Intent} \mid \text{Sexist}, \mathbf{x})$};
\node[var_node] (t23) [below right=2.2cm and 0.5cm of t21] {\textbf{Task 2.3}\\ $\hat{p}^{(3)}_c = P(\text{Category}_c \mid \text{Sexist}, \mathbf{x})$};

\node[op_node] (mul2) [above=0.5cm of t22] {$\otimes$};
\node[op_node] (mul3) [above=0.5cm of t23] {$\otimes$};


\draw[->] (gemini) -- (mlp);

\draw[->] (mlp) -- (t21) node[midway, right] {$\mathbf{h}$};

\draw[->, rounded corners] (mlp.west) -| (mul2.north) node[pos=0.4, above] {$\mathbf{h}$};
\draw[->, rounded corners] (mlp.east) -| (mul3.north) node[pos=0.4, above] {$\mathbf{h}$};

\draw[->, dashed] (t21.west) -| (mul2.north) node[pos=0.25, above] {$\hat{p}^{(1)}$};
\draw[->, dashed] (t21.east) -| (mul3.north) node[pos=0.25, above] {$\hat{p}^{(1)}$};

\draw[->] (mul2) -- (t22);
\draw[->] (mul3) -- (t23);

\end{tikzpicture}
\end{adjustbox}
\caption{Architectural data flow illustrating the hierarchical conditional dependencies. Pre-trained Gemini 2 embeddings are mapped to a shared semantic representation $\mathbf{h}$ via SwiGLU blocks. This shared representation is routed to all classification heads (solid lines). The dashed lines and multiplication nodes ($\otimes$) represent the conditional dependency: structurally enforced via detached soft-gating in Run 3, and probabilistically enforced across all runs via joint probability decoding ($P(\text{Subtask}) = \hat{p}^{(1)} \cdot \hat{p}^{(\text{sub})}$) and masked multi-task loss.}
\label{fig:arch_overview}
\end{figure}
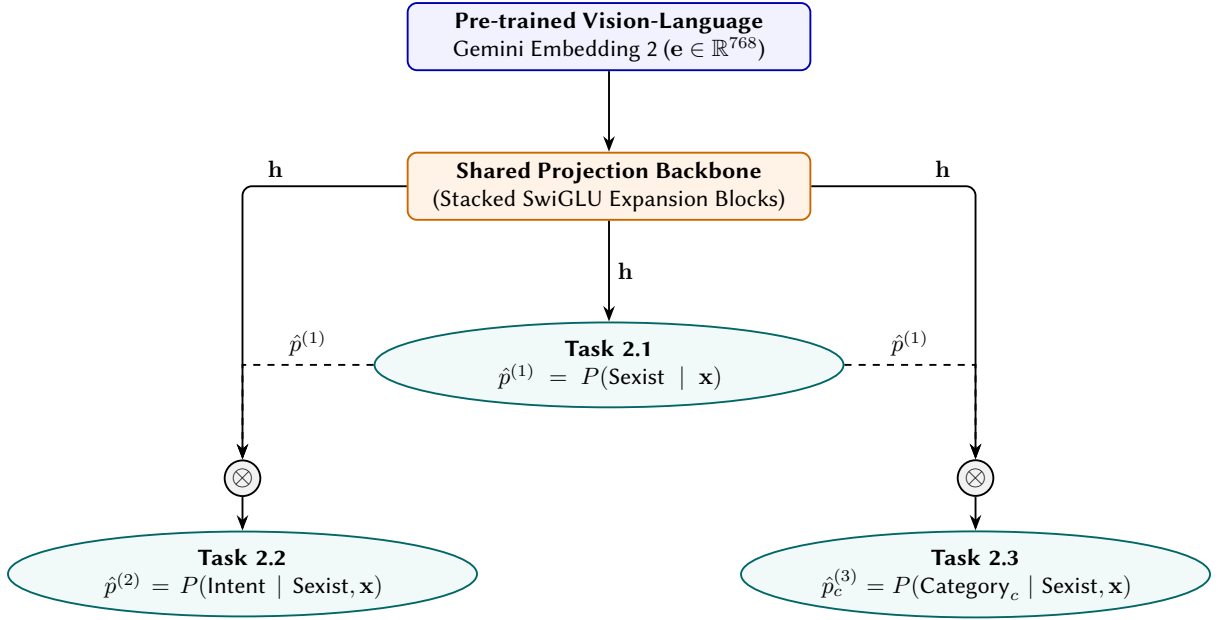

\subsection{Soft-Label Training and Dynamic Task Weighting}
\label{sec:kl}

The Soft-Soft evaluation expects systems to predict full annotator distributions. We train each head against empirical soft targets using KL divergence~\cite{kullback1951information}, which penalizes the model when it places mass on categories annotators collectively rejected. For Task~2.3, binary KL is applied independently per category and averaged.

Balancing three heterogeneous tasks—differing in label spaces, class imbalance, conditional support, and intrinsic difficulty—poses an optimization challenge. Fixed scalar weights require extensive tuning and cannot adapt as tasks converge at different rates. We instead adopt homoscedastic uncertainty weighting~\cite{kendall_gal_cipolla_2018}, introducing learnable variance parameters $\sigma_i$:

\begin{equation}
    \mathcal{L}(\mathbf{W}, \sigma_1, \sigma_2, \sigma_3) = \sum_{i=1}^{3} \left( \frac{1}{2\sigma_i^2}\mathcal{L}_i + \log \sigma_i \right).
\end{equation}

This formulation enables self-paced learning: when task losses are high, $\sigma_i$ increases to reduce that task's contribution; the $\log \sigma_i$ term prevents complete task abandonment. Finally, losses for Tasks~2.2 and~2.3 are masked to propagate gradients only for instances where downstream annotations exist, preventing non-sexist memes from forcing arbitrary updates on conditional heads.

\subsection{Detached Soft-Gating (Ablation)}

As an architectural ablation (submitted as Run 3), we tested explicit structural conditioning: before downstream heads, the shared representation is scaled by the predicted sexism probability with stopped gradients:

\begin{equation}
    \mathbf{h}_{\mathrm{downstream}} = \mathbf{h} \odot \operatorname{sg}\left[\hat{p}^{(1)}\right].
\end{equation}

This suppresses downstream features when sexism probability is low. The stop-gradient operator prevents downstream losses from corrupting the Task~2.1 detector. The comparable performance of this variant (Section~\ref{sec:results}) suggests hierarchical constraints are adequately captured through conditional loss masking and probabilistic decoding alone, without architectural gating.

\section{Training and Inference}

We split training data into 80/10/10\% partitions stratified by Task~2.1 majority label. Models are optimized with AdamW (learning rate $10^{-4}$, weight decay $10^{-2}$, batch size 8), OneCycleLR scheduling with 30\% warm-up and cosine annealing, \texttt{bf16-mixed} precision, and early stopping on validation loss. We submitted three runs: two base Gated MLP models with different random seeds (Runs 1–2) and the detached soft-gating variant (Run 3).

\subsection{Inference}

\textbf{Soft-label decoding.} For Task~2.2, the joint distribution over \{\texttt{JUDGEMENTAL}, \texttt{DIRECT}, \texttt{NO}\} is:
\begin{equation}
    \left\{\hat{p}^{(1)}\hat{p}^{(2)},\ \hat{p}^{(1)}(1-\hat{p}^{(2)}),\ 1-\hat{p}^{(1)}\right\}.
\end{equation}
For Task~2.3, each category probability is $\hat{p}^{(1)}\hat{p}^{(3)}_c$, with $1-\hat{p}^{(1)}$ assigned to the non-sexist complement.

\textbf{Hard-label decoding} (submitted to Hard-Hard track without threshold tuning). We threshold at $0.5$: if $\hat{p}^{(1)} < 0.5$, all downstream labels are null. Otherwise, Task~2.2 uses $\hat{p}^{(2)} \geq 0.5$. For Task~2.3, the active set is $\{c \mid \hat{p}^{(3)}_c \geq 0.5\}$. To satisfy the ontological constraint that sexist memes must have $\geq 1$ category, empty sets are resolved by selecting the maximum-probability category. No task-specific tuning was performed.

\section{Results}
\label{sec:results}

\subsection{Local Evaluation}

Tables~\ref{tab:internal_eval_all_soft} and~\ref{tab:internal_eval_all_hard} report results on our 10\% test split using official ICM metrics~\cite{amigo2022evaluating}, with cross-entropy (CE) included for binary tasks.

\begin{table}[htpb]
    \centering
    \caption{Soft-label evaluation on local test split.}
    \begin{adjustbox}{max width=\linewidth}
    \begin{tabular}{l|ccc|ccc|cc}
    \toprule
    & \multicolumn{3}{c|}{Task 2.1} & \multicolumn{3}{c|}{Task 2.2} & \multicolumn{2}{c}{Task 2.3} \\
    Model
    & ICM-S & ICM-S Nr & CE
    & ICM-S & ICM-S Nr & CE
    & ICM-S & ICM-S Nr \\
    \midrule
    Run 1 (Base) &
    0.1648 & 0.5258 & 0.9340
    & \textbf{-0.6040} & \textbf{0.4367}
    & 1.5011 & -3.6935 & 0.3111 \\
    Run 2 (Base) &
    \textbf{0.2025} & \textbf{0.5317} & \textbf{0.9314}
    & -0.7807 & 0.4181 & \textbf{1.4822}
    & \textbf{-3.2130} & \textbf{0.3356} \\
    Run 3 (Gating) &
    0.1900 & 0.5297 & 0.9448
    & -0.8484 & 0.4110 & 1.5426
    & -3.2546 & 0.3335 \\
    \bottomrule
    \end{tabular}
    \end{adjustbox}
    \label{tab:internal_eval_all_soft}
\end{table}

\begin{table}[htpb]
    \centering
    \caption{Hard-label evaluation on local test split. F1Y = F1 for \texttt{YES}, M F1 = macro F1.}
    \begin{adjustbox}{max width=\linewidth}
    \begin{tabular}{l|ccc|ccc|ccc}
    \toprule
    & \multicolumn{3}{c|}{Task 2.1} & \multicolumn{3}{c|}{Task 2.2} & \multicolumn{3}{c}{Task 2.3} \\
    Model
    & ICM-H & ICM-H Nr & F1Y
    & ICM-H & ICM-H Nr & M F1
    & ICM-H & ICM-H Nr & M F1 \\
    \midrule
    Run 1 (Base) &
    \textbf{0.3308} & \textbf{0.6690} & \textbf{0.7524} &
    \textbf{0.1295} & \textbf{0.5445} & \textbf{0.5189} &
    -0.4355 & 0.4098 & 0.4045 \\
    Run 2 (Base) &
    0.2635 & 0.6346 & 0.7407 &
    -0.1270 & 0.4563 & 0.4886 &
    \textbf{-0.3319} & \textbf{0.4313} & \textbf{0.4530} \\
    Run 3 (Gating) &
    0.2917 & 0.6490 & 0.7488 &
    -0.0476 & 0.4836 & 0.4816 &
    -0.4213 & 0.4128 & 0.4292 \\
    \bottomrule
    \end{tabular}
    \end{adjustbox}
    \label{tab:internal_eval_all_hard}
\end{table}

Local results reveal meaningful seed variance, especially for downstream tasks—consistent with the small-data, high-disagreement setting. Run~2 performed best for Task~2.3, while Run~1 led for Task~2.2. The gating variant (Run 3) performed comparably to the base architecture, confirming that hierarchical constraints are effectively captured through conditional losses and probabilistic decoding without architectural modification.

\subsection{Official Leaderboard}

Table~\ref{tab:official_soft} reports our best official Soft-Soft results per task. Our system ranked \textbf{1st} on Task~2.3 and \textbf{4th} on Tasks~2.1 and~2.2.

\begin{table}[htpb]
\centering
\caption{Official Soft-Soft leaderboard results.}
\label{tab:official_soft}
\begin{tabular}{l l c c c}
\toprule
Task & System & Rank & ICM-Soft & ICM-Soft Norm \\
\midrule
\textbf{2.1} & \textbf{aiwizards\_3} & \textbf{4} & \textbf{0.2323} & \textbf{0.5373} \\
2.1 & aiwizards\_1 & 5 & 0.2039 & 0.5328 \\
2.1 & aiwizards\_2 & 6 & 0.1846 & 0.5297 \\
\midrule
\textbf{2.2} & \textbf{aiwizards\_1} & \textbf{4} & \textbf{-0.6720} & \textbf{0.4285} \\
2.2 & aiwizards\_3 & 5 & -0.7623 & 0.4189 \\
2.2 & aiwizards\_2 & 7 & -0.9496 & 0.3990 \\
\midrule
\textbf{2.3} & \textbf{aiwizards\_2} & \textbf{1} & \textbf{-2.8881} & \textbf{0.3469} \\
2.3 & aiwizards\_3 & 2 & -2.9507 & 0.3436 \\
2.3 & aiwizards\_1 & 3 & -3.2537 & 0.3276 \\
\bottomrule
\end{tabular}
\end{table}

Table~\ref{tab:official_hard} reports our best Hard-Hard results, obtained without threshold tuning or hard-label-specific calibration. The gap between Soft-Soft and Hard-Hard rankings confirms that distribution prediction and discrete decision quality are related but distinct objectives, with the latter likely benefiting from task-specific threshold optimization.

\begin{table}[htpb]
\centering
\caption{Official Hard-Hard leaderboard results.}
\label{tab:official_hard}
\begin{tabular}{l l c c c}
\toprule
Task & System & Rank & ICM-Hard & ICM-Hard Norm \\
\midrule
\textbf{2.1} & \textbf{aiwizards\_3} & \textbf{14} & \textbf{0.2673} & \textbf{0.6359} \\
2.1 & aiwizards\_1 & 18 & 0.2208 & 0.6123 \\
2.1 & aiwizards\_2 & 19 & 0.2188 & 0.6112 \\
\midrule
\textbf{2.2 }& \textbf{aiwizards\_3} & \textbf{12} & \textbf{0.0020} & \textbf{0.5007} \\
2.2 & aiwizards\_1 & 13 & -0.0353 & 0.4877 \\
2.2 & aiwizards\_2 & 17 & -0.1248 & 0.4566 \\
\midrule
\textbf{2.3} & \textbf{aiwizards\_3 }& \textbf{8} & \textbf{-0.3953} & \textbf{0.4180} \\
2.3 & aiwizards\_2 & 9 & -0.4224 & 0.4124 \\
2.3 & aiwizards\_1 & 13 & -0.5316 & 0.3897 \\
\bottomrule
\end{tabular}
\end{table}

Three observations emerge: (1) the system is better matched to soft-label prediction, for which it was designed; (2) performance is strongest on the most challenging subtask (Task~2.3); (3) probabilistic outputs retain discriminative structure sufficient for non-trivial hard-label results even with naive thresholding.

\section{Limitations}

Our methodology is subject to three principal limitations. First, we rely on Gemini Embedding~2, a proprietary model, which constrains the full reproducibility of the proposed pipeline. However, the modular architecture of our system allows straightforward substitution with open-weight alternatives in future work. Second, the hard-label decoding procedure employs a fixed decision threshold of 0.5; consequently, the Hard-Hard results should not be construed as representing an upper bound on attainable performance. Third, our statistical analysis of physiological signal separability was confined to EEG data; gaze and heart-rate signals were not empirically evaluated and were excluded a priori, based on considerations of feature dimensionality and hypothesized task relevance, rather than on evidence derived from measured separability. Future work might include EEG, eye-tracking, and heart-rate data as well as their combination~\cite{gabaldn-etal-2026-human, alacam-etal-2024-eyes, oguz-emotion-2023, 10636286}.

\section{Conclusion}

We presented a hierarchical multi-task system for soft-label sexism detection in memes. By combining frozen Gemini Embedding~2 representations, compact gated MLP blocks, KL divergence against annotator distributions, and learned homoscedastic uncertainty weighting, our approach achieves strong performance while maintaining efficiency (3.5M trainable parameters). Our submissions ranked 1st on Task~2.3 and 4th on Tasks~2.1 and~2.2 on the official Soft-Soft leaderboards. Future work should investigate open-weight embedding backbones, task-specific hard-label calibration, and deeper integration of physiological signals through non-linear fusion architectures.

\section*{Declaration on Generative AI}
\noindent
During the preparation of this work, the authors used OpenAI-ChatGPT in order to: Grammar and spelling check, Paraphrase and reword. After using this tool/service, the authors reviewed and edited the content as needed and assume full responsibility for the content of the publication.

\begin{acknowledgments}
Luca Tedeschini was supported by the project ``IPCEI-CIS - Progetto Villanova'' (Beneficiary: Tiscali Italia S.p.A., Prog. n. SA. 102519 -- CUP B29J24000850005).
\end{acknowledgments}

\bibliography{main}
\end{document}